 \documentclass[final,12pt]{include/clear2023}
\usepackage{caption}
\usepackage{wrapfig}
\usepackage{algorithm}
\usepackage{algorithmic}
\usepackage{mathtools}

\usepackage[utf8]{inputenc} %
\usepackage[T1]{fontenc}    %
\usepackage{hyperref}       %
\usepackage{url}            %
\usepackage{booktabs}       %
\usepackage{amsfonts}       %
\usepackage{nicefrac}       %
\usepackage{microtype}      %
\usepackage{xcolor}         %

\usepackage{times}
\usepackage{cleveref}
\usepackage{titletoc}
\usepackage{appendix}
\crefname{appsec}{appendix}{appendices}
\Crefname{appsec}{Appendix}{Appendices}

\definecolor{mydarkblue}{rgb}{0,0.08,0.45}
\hypersetup{ %
    pdftitle={},
    pdfauthor={},
    pdfsubject={},
    pdfkeywords={},
    pdfborder=0 0 0,
    pdfpagemode=UseNone,
    colorlinks=true,
    linkcolor=mydarkblue,
    citecolor=mydarkblue,
    filecolor=mydarkblue,
    urlcolor=mydarkblue,
    pdfview=FitH
}

\title[Can Active Sampling Reduce Causal Confusion in Offline RL?]{Can Active Sampling Reduce Causal Confusion in\\Offline Reinforcement Learning?}

\clearauthor{%
 \Name{Gunshi Gupta} \addr{University of Oxford} \Email{gunshi.gupta@lmh.ox.ac.uk}
 \\
 \Name{Tim G. J. Rudner} \addr{University of Oxford}
  \\
 \Name{Rowan McAllister} \addr{Toyota Research Institute, USA}
  \\
 \Name{Adrien Gaidon} \addr{Toyota Research Institute, USA}
  \\
 \Name{Yarin Gal} \addr{University of Oxford}
}

\begin{document}

\maketitle

\begin{abstract}
Causal confusion is a phenomenon where an agent learns a policy that reflects imperfect spurious correlations in the data. Such a policy may falsely appear to be optimal during training if most of the training data contain such spurious correlations. This phenomenon is particularly pronounced in domains such as robotics, with potentially large gaps between the open- and closed-loop performance of an agent. In such settings, causally confused models may appear to perform well according to open-loop metrics during training but fail catastrophically when deployed in the real world. In this paper, we study causal confusion in offline reinforcement learning. We investigate whether selectively sampling appropriate points from a dataset of demonstrations may enable offline reinforcement learning agents to disambiguate the underlying causal mechanisms of the environment, alleviate causal confusion in offline reinforcement learning, and produce a safer model for deployment. To answer this question, we consider a set of tailored offline reinforcement learning datasets that exhibit causal ambiguity and assess the ability of active sampling techniques to reduce causal confusion at evaluation. We provide empirical evidence that uniform and active sampling techniques are able to consistently reduce causal confusion as training progresses and that active sampling is able to do so significantly more efficiently than uniform sampling. Our code is available at \url{https://github.com/gunshi/offline_active_rl}.
\end{abstract}


\section{Introduction} \label{sec:intro}

Offline learning offers opportunities to scale reinforcement learning to domains where offline data is plentiful but online interaction with the environment is costly.
The fundamental challenge of offline reinforcement learning is to identify the cause and effect of actions from a fixed dataset, which is often intractable.
In the absence of online interactions, our hope is that the dataset uniformly covers an exhaustive set of plausible scenarios.
This is often not the case in datasets for robotic control, which are heavy-tailed and often contain only a handful of samples for rare (and informative) events.
Causal confusion occurs when agents misinterpret the underlying causal mechanisms of the environment and erroneously associate certain actions or states with a given reward.
For example, if an agent happens to simultaneously observe independent events $X$ and $Y$ in its environment whenever it receives a reward $R$, and $R$ causally depends on $Y$ but not on $X$, the agent may attribute the reward $R$ to $X$ and $Y$ occurring jointly even though $R$ may be independent of $X$.
Problematically, if the spurious correlation between $X$ and $R$ observed at training time ceases to hold at deployment time, a causally-confused model may show a significant deterioration in performance.

Although spurious features might not \emph{perfectly} explain all the input-label pairs in offline data, optimisation methods such as stochastic gradient descent benefit from correlations in the data when seeking to reduce the training loss. Therefore spurious correlations are unwittingly transferred from the data to the mechanisms of learned models.
This phenomenon is especially pronounced in models trained on high-dimensional visual inputs, since extracting the true causal factors of an environment (and their interplay) from an image is a particularly difficult problem.
On the other hand, it is noticeably easy for neural networks to find shortcuts for prediction, for instance, using co-occurrence patterns between objects and backgrounds to successfully perform object detection, as has been extensively documented in the deep computer vision literature~\citep{terra, shortcut}. Several works have also reported causal confusion in control policy learning and used heuristic loss re-weighting or dataset-balancing schemes based on domain knowledge of the prediction task to reduce the amount of causal confusion in the learned models~\citep{eqa_matterport,sergey_offline_rl}.
However, these heuristics require a practitioner to know the source of causal confusion \emph{a priori} since they explicitly upweigh the loss for specific spurious-correlation-breaking samples in the dataset~\citep{keyframe, copycat}.
While this may be limiting in practice, the success of these heuristics demonstrates that it is possible to recover less causally-confused models from a fixed dataset, as long as we have access to an \emph{oracle} sampling scheme for the training data.

This insight suggests that causal confusion in learned models may primarily be an artifact of the optimisation procedure and raises the question of whether systematically providing data points that expose spurious correlations to the optimizer may allow alleviating causal confusion.
Unfortunately, in practice, identifying such sets of data points for a given task is challenging.
Furthermore, naive sampling schemes, such as uniform sampling from the dataset for a sufficiently long training horizon, may eventually be able to resolve causal confusion in a learned model.
For instance, if a model has achieved close-to-zero loss on the dominant scenarios, high loss from the tail-cases may begin to influence optimisation.
However, even if such naive approaches were successful at partially or fully resolving causal confusion, for reasons of computational efficiency and predictive performance (e.g., early stopping), their practical usefulness would be limited.

In this paper, we investigate the efficacy of data sampling strategies in mitigating causal confusion in offline reinforcement learning from datasets exhibiting causal ambiguity.
Our contributions can be summarised as follows:\vspace*{-3pt}
\begin{enumerate}
\setlength\itemsep{-3pt}
    \item We design three decision-making tasks to study causal confusion in offline RL from visual inputs:
a grid-world simulation of traffic-light navigation; a maze navigation task with goals correlated to a fixed position in the maze; and a car-racing game with highly-correlated action sequences across time.
    \item We study uncertainty-based and loss-based active sampling techniques and find that active sampling alleviates causal confusion in offline training and yields a higher reward at evaluation.
    \item We further show that active sampling is able to alleviate causal confusion at a significantly higher sample efficiency than naive uniform sampling, and that the usefulness of active sampling in alleviating causal confusion is highly related to the quality of predictive uncertainty estimates used in the best-performing, uncertainty-based acquisition function.
\end{enumerate}
Our work suggests that in many cases we may not have saturated our datasets and can push the capabilities of offline reinforcement learning agents by optimising on the long tail of the data.
We show that simply showing the same training dataset according to a better distribution to a model during training could be a viable strategy to explore before investing efforts in collecting more high-quality data or exploring larger models.\vspace*{-10pt}
\section{Preliminaries}
\label{prelims}

We consider an environment formulated as a Markov Decision Process (MDP) $\mathcal{M}$ defined by a tuple $\left(\mathcal{S}, \mathcal{A}, \mathcal{P}, r, d_0, \gamma\right)$, where $\mathcal{S}$ is the state space, $\mathcal{A}$ is the action space, $\mathcal{P}\left(\mathbf{s}^{\prime} \mid \mathbf{s}, \mathbf{a}\right)$ is the transition probability distribution, $r: \mathcal{S} \times \mathcal{A} \rightarrow \mathbb{R}$ is the reward function, $d_0$ is the initial state distribution, and $\gamma \in(0,1]$ is the discount factor. The goal of reinforcement learning (RL) is to find an optimal policy $\pi(\mathbf{a} \mid \mathbf{s})$ that maximizes the cumulative discounted reward $\mathbb{E}_{\mathbf{s}_t, \mathbf{a}_t}\left[\sum_{t=0}^{\infty} \gamma^t r\left(\mathbf{s}_t, \mathbf{a}_t\right)\right]$, where $\mathbf{s}_0 \sim d_0(\cdot), \mathbf{a}_t \sim \pi\left(\cdot \mid \mathbf{s}_t\right)$, and $\mathbf{s}_{t+1} \sim \mathcal{P}\left(\cdot \mid \mathbf{s}_t, \mathbf{a}_t\right)$.

$Q$-learning-based RL algorithms learn an optimal state--action value function $Q^*(s, a)$, representing the expected cumulative discounted reward starting from $s$ with action $a$ and then acting optimally according to policy $\pi^*$ thereafter, ie $Q^*(s, a) = \mathbb{E}_{\pi^*}\left[\sum_{t=0}^{\infty} \gamma^t r\left(\mathbf{s}_t, \mathbf{a}_t\right) \mid s_0=s, a_t=a\right]$.
Analogously, the value function $V(s) = \mathbb{E}_{\pi}\left[\sum_{t=0}^{\infty} \gamma^t r\left(\mathbf{s}_t, \mathbf{a}_t\right) \mid s_0=s\right]$ represents the expected cumulative discounted reward achievable from state $s$ when following policy $\pi$.
$Q$-learning is trained on the Bellman equation defined as follows with the Bellman optimal operator $\mathcal{B}$ defined by:
\begin{equation}
\label{bellman_update}
    \mathcal{B} Q(s, a):=R(s, a)+\gamma \mathbb{E}_{P\left(s^{\prime} \mid s, a\right)}\left[\max _{a^{\prime}} Q\left(s^{\prime}, a^{\prime}\right)\right].
\end{equation} 
The $Q$-function is updated by minimizing the Bellman Squared Error $\mathbb{E}\left[(Q-\mathcal{B} Q)^2\right]$ where a frozen, periodically updated copy of the $Q$-network weights are used to compute the target $\mathcal{B}Q$. 

Offline RL algorithms aim to learn an optimal policy by learning estimates of the value (or $Q$-value) function from a static dataset of transitions $\mathcal{D}=\left\{\left(\mathbf{s}, \mathbf{a}, r, \mathbf{s}^{\prime}\right)\right\}$ collected by a behaviour policy $\pi_\beta$.\
Since the agent does not interact with the environment, naively using online RL techniques in the offline case leads to value function overestimation on unseen states and actions.
This happens because the \emph{max} operator in the target update in~\Cref{bellman_update} propagates erroneously high values from the next state's value estimate into the update. 
Thus, algorithms based on the \emph{pessimism principle} of underestimating the $Q$-values to optimise the worst-case regret bound have been successful at learning policies from datasets containing either good coverage of the state--action space or high return trajectories~\citep{cql, bear, buckman2021the, pbrl}.

Prior work categorised pessimistic offline RL algorithms into (1) \emph{proximal} and (2) \emph{uncertainty-aware} algorithms~\citep{buckman2021the}. The former penalises action-value estimates based on deviations from the actions seen in the dataset, while the latter conservatively updates the value functions, taking the uncertainty of their targets into account. 
Proximal pessimistic algorithms like CQL, BEAR ~\citep{bear,cql} are known to work well with exactly the kind of narrow and biased data distributions that are most prone to causal confusion ~\citep{d4rl}.
We, therefore, analyse the proximal class of algorithms in this work and leave the study of causal confusion in uncertainty-aware pessimistic offline RL algorithms to future work.

\paragraph{Conservative $Q$-Learning.}
We choose CQL~\citep{cql} as an instantiation of a proximal pessimistic offline RL algorithm in our experiments, due to its simplicity and competitive performance on offline RL benchmarks.
The CQL objective,  which combines the standard TD-error of $Q$-learning with a penalty constraining deviations from the behaviour policy, is defined as: 

\begin{align}
\SwapAboveDisplaySkip
    \mathcal{L}_{\text {critic }}^{\text {CQL }}(\theta)
    \hspace*{-2pt}=\hspace*{-2pt}
    \frac{1}{2} \underset{\left(s, a, s^{\prime}\right) \sim \mathcal{D}}{\mathbb{E}}\left[\left(Q_{\theta}(s,a)-\mathcal{B} Q_{\bar{\theta}}(s,a)\right)^2\right]
    \hspace*{-2pt}+\hspace*{-2pt}
    \alpha_0 \underset{s \sim \mathcal{D}}{\mathbb{E}}\hspace*{-2pt}\left[\log \sum_a \exp Q_\theta(s, a)
    \hspace*{-2pt}-\hspace*{-2pt}
    \underset{a \sim {\pi}_\beta}{\mathbb{E}}[Q_\theta(s, a)]\right],
    \label{eqn:cql_objective}
\end{align}
where $\alpha_0$ is a coefficient controlling the degree of conservatism.

\section{Related Work}
\label{sec:related_work}

\paragraph{Causal Confusion in Supervised Policy Learning.}
Several works in offline imitation learning have proposed solutions to mitigate causal confusion. 
\citet{dehaan} demonstrated causal mis-identification in models trained on expert trajectories collected in the Mountain-Car and CARLA simulators ~\citep{sutton2018reinforcement, carla} where inputs were augmented with the previous control command taken by the acting agent.
They proposed to resolve the confusion through a scheme to query an expert or collect additional rollouts in the environment to refine a learned causal graph that conditions a learned policy.
\citet{copycat} propose adversarial training to prune out any \emph{known} sources of spurious correlations from the policy's representation, for instance, the previous control commands given to a robot;~\citet{keyframe} propose re-weighting the losses of data points based on the loss of a model trained with just the spurious correlates as the input;
OREO~\citep{oreoneurips} regularises the model's representation to be invariant to any individual object being dropped out in a scene.  
\citet{lee2022divdis} propose training a diversified policy ensemble for imitation learning in the case when \emph{perfect spurious correlations} exist in the data and later select from these hypotheses based on validation data
Causal Confusion has also recently been studied in reward-learning from preferences~\citep{rewardlearning}, where spurious correlations can be drawn between a human evaluator's preferences and certain actions or parts of the state space, for tasks in the Assistive gym~\citep{assistive}. 
For instance, a reward model trained to classify or rank trajectories for a feeding task can easily learn that higher force correlates to higher reward since a certain level of force needs to be inserted in the direction of the mouth in all preferred trajectories.

\paragraph{Active Sampling.}
The analysis in this work is inspired by the approach presented in~\citet{causal_bald}, which considers the problem of treatment effect estimation in settings where we wish to be sample-efficient in terms of querying for outcomes of costly experiments.
\citet{causal_bald} propose several causality-inspired acquisition functions that prefer data points that have both high variance in their estimated outcomes and that correspond to covariates with considerable overlap in the dataset.
In $Q$-learning-based RL algorithms, Prioritised Experience Replay~\citep{per} is a non-Bayesian loss-based sampling scheme proposed for off-policy learning.
It computes acquisition scores based on the TD-error of transitions and has not been studied in offline RL.

\paragraph{Ensemble Models in RL.}
Ensembles have been studied extensively to guide exploration in online RL~\citep{ucb_dqn_exploration,sunrise}, and recently to construct adaptive pessimism constraints in offline RL, to disincentivise uncertain actions from having high estimated returns. 
It was recently also shown that significantly increasing the size and diversity of the ensembled critic in Soft-Actor-Critic~\citep{sac} performs competitively with state-of-the-art offline RL algorithms~\citep{edac}. 
We are not aware of any prior work that has explored how uncertainty about the value function ca be used to sample state transitions in RL.

\paragraph{AI Alignment.}
AI alignment seeks to align the behavior of agents with the intentions of their creators by investigating the incentives behind demonstrated tasks.
Recent work on \emph{Goal Mis-generalisation}~\citep{goalmisgen} explores how online RL agents in Procgen~\citep{cobbe2019procgen} can get confused about the goal they are pursuing since those goals co-occur with irrelevant artifacts in the environment most of the time.
In this case, the specification is correct, but the agent still pursues an unintended objective (as opposed to poor reward definitions that predictably lead to reward hacking).
\section{Alleviating Causal Confusion in Offline RL via Active Sampling}
\label{sec:method}

In this section, 
we provide relevant background on active sampling and describe the active sampling schemes we use in this work in conjunction with offline RL.

\subsection{Active Sampling}
\label{active sampling}

Active sampling techniques are used to selectively sample from a given dataset during training to enable sample-efficient learning or to improve learning from noisy data~\citep{online_batch,coresets}.
To perform active sampling, data points are scored according on a given acquisition function, and a small number of points are sampled based on these scores using a pre-specified weighting scheme.
Similar to active learning, the acquisition functions are often information-theoretic quantities and seek to reduce the uncertainty about model parameters by acquiring informative data points.
In this work, we build on~\citet{causal_bald} and study whether active sampling can enable sample-efficient resolution of causal confusion in models trained on long-tailed datasets.
In particular, the focus of this work is to investigate whether active sampling techniques can alleviate the negative effects of causal ambiguity in state--action trajectories used in offline RL---without necessitating any modifications to the learning objective.
\Cref{alg:practical_alg_active} describes the procedure for CQL, integrated with our proposed method for active sampling of state transitions.
The modifications from uniform sampling are highlighted in blue.
As we noted in \Cref{sec:intro}, causal confusion in policy learning has been observed to occur due to insufficient exposure of a model to the rare scenarios that make up the long tail of the dataset it is trained on.
These data points are unknown in advance, and we hypothesise that a good approach to finding them is to compute the epistemic uncertainty or the loss of the training model of each point in the dataset.
We will thus define a set of loss-based and uncertainty-based acquisition functions, which we will use to perform active sampling, next.

\paragraph{Uncertainty-based Sampling.}
We are interested in sampling state transitions for which the learned $Q$-network is uncertain about its predictions over the action seen in the data ($a_{\beta} = \pi_{\beta}(s)$) or its own greedy action ($a^* = \arg \max Q(s)$).
We model the epistemic uncertainty in the learned $Q$-function by creating an ensemble model of $Q$-functions $\{ Q_{\theta^{i}} \}_{i}$ and training each of them on identical transitions across the ensemble members, with their own corresponding targets, $\{ \bar{\theta^{i}} \}_{i}$, as proposed in~\citet{ghasemipour2022msg}.
We can then use the variance of the $Q$-value estimates of the ensemble as a measure of the epistemic uncertainty.
However, the $Q$-values of different ensemble members may have arbitrary numerical offsets (and still be equivalent) since they are trained by bootstrapping their own value estimates.
To address this, we estimate the uncertainty about an action's value by computing the variance of its advantage estimates over the ensemble, where the advantage of action $a^{i}$ for a $Q$-learner is given by
\begin{align}
\SwapAboveDisplaySkip
    A^{\pi}(s, a^{i})=Q^{\pi}(s, a^{i})-V^{\pi}(s) \approx Q^{\pi}(s, a^{i}) - \sum_{a} \left[Q(s,a) \cdot \frac{e^{Q(s, a)}}{\sum_{a'} e^{Q(s, a')}}\right].
\end{align}
The advantage function in RL represents a causal quantity assessing the relative \emph{effect} of action $a^i$ on the outcome $Q$ for a given state $s$. 
The acquisition scores will then be the variance of the advantage estimates: $\textrm{Var}(A^{\pi}(s, a^{i}))$.
We consider two variants, \emph{Variance-greedy} and \emph{Variance-data}, referring to whether the advantages are computed for the greedy actions ($a^i=a^*$) or the dataset actions ($a^i=a^{\beta}$).
To ensure a fair comparison, we will fix the ensemble size we use for the $Q$-network across all the sampling schemes we try, including uniform sampling.

\paragraph{Loss-based Sampling.}
We sample transitions based on their Temporal Difference error similar to Prioritised Experience Replay (PER; ~\citet{per}) and refer to this variant as \emph{TD-Error}.

We discuss a causal interpretation of our active sampling approach in \Cref{cate-connections} and explain how it relates to causal discovery of accurate treatment effects from offline data.

\begin{wrapfigure}{r}{0.5\textwidth}
\begin{small}
\vspace*{-17pt}
\begin{minipage}[t]{0.99\linewidth}
\begin{algorithm}[H]
\small
\caption{
    Conservative $Q$-Learning ( + active sampling)
}
\label{alg:practical_alg_active}
\begin{algorithmic}[1]
    \STATE Initialise ensemble $Q$-function $Q_{\theta}$, $n_{ep}$=epochs, $d_{sz}$=dataset size, $b_{sz}$=batch size, $T$=steps-per-epoch.

    \FOR{epoch $e$ in \{1, \dots, $n_{ep}$\}}

        \FOR{step $t$ in \{1, \dots, T\}}
            \STATE  \textcolor{blue} {compute scores $acq_i$ over $\mathcal{D}_{\text {train }}=[s_i, a_i]_{i=1}^{d_{sz}}$ according to the acquisition functions in \Cref{active sampling}}
            \STATE  \textcolor{blue} { $acq_i = \frac{acq_i}{\sum_{j=1}^{d_{sz}} acq_j}$ (normalise scores)}
            \STATE  \textcolor{blue}{sample batch $B=[s_i, a_i, s'_i, r_i]_{i=1}^{b_{sz}}$ from $\mathcal{D}_{\text {train }}$  $\sim multinomial (acq)$}

            \STATE Train the $Q$-function on $D_{train}$ using objective from ~\Cref{eqn:cql_objective} \\
        \ENDFOR
    \ENDFOR

\end{algorithmic}
\end{algorithm}
\end{minipage}
\end{small}
\end{wrapfigure}

\subsection{Computing Acquisition Scores}
\label{computing}

In practice, computing the acquisition scores over all the transitions in the dataset can be both expensive and redundant since high-error or high-uncertainty points will likely stay high over a short window of subsequent gradient steps. 
Thus, we only recompute all the scores after every $n$ gradient steps and vary $n$ as a hyper-parameter in our experiments.

Later in~\Cref{ale-results}, we will also explore a scheme where scores are recomputed only over sampled batches, as done in Prioritised Experience Replay (PER). 
This is an approximation where a priority queue is maintained and the priority of every data point is derived from the TD-Error computed on it. In this case, the priorities are only updated for a small subset of points at every gradient step, since the acquisition scores are only recomputed on the data points in the sampled batch, potentially leading to many points in the replay buffer with \emph{stale} scores. 
We adopt the same implementation as that of PER with the scores for the priorities coming from the TD-error and Variance estimates. We include further details about this procedure in the appendix.
We refer to the above two cases by appending \emph{-dataset} and \emph{-batch} to the names of the sampling schemes to indicate that scores are recomputed for the entire data versus just the sampled batch respectively.

\section{Experiments}
\label{experiments}

In this section, we first describe the experimental setup and evaluation protocol (in \Cref{expt-design}).
Next, we introduce the different benchmark domains and present the results of the corresponding empirical evaluations (in \Cref{traffic-results,procgen-results,ale-results,uncertainty-quality-results}).

\subsection{Experiment Design}\label{expt-design}
We focus on investigating the following questions through our experiments:
\textbf{(1)} Can causal confusion be consistently observed in offline RL agents when sampling transitions uniformly from a long-tailed demonstration dataset?
\textbf{(2)} Can active sampling based on a model's predictive uncertainty, or its loss, help in alleviating the effects of causal confusion?
\textbf{(3)} To what extent does the quality of predictive uncertainty estimates affect predictive performance and sample efficiency gains under uncertainty-based active sampling?

To answer these questions, we modify three reinforcement learning environments (MiniGrid, Atari, and Procgen) to either include a realistic spurious correlate in the observations or bias the sampling of episodes.
The specific modifications ensure that a demonstration dataset collected in these environments produces causally-confused agents when trained with an offline RL algorithm with uniform sampling of the data.
The modifications are described and motivated further in the following sections.

Prior work on causal confusion, largely in the domain of imitation learning, has used a training protocol where training is terminated once a model reaches a sufficiently low loss \citep{dehaan}.
The last checkpoint is then evaluated in the environment to report the achieved reward.
This procedure, however, does not capture the full picture and disregards certain practical considerations of training neural networks---especially in the offline RL setting---for the following two reasons:
\begin{enumerate}
    \item This approach cannot be adopted to offline RL as offline RL training does not lead to monotonically increasing performance, and the Q-function starts to excessively optimise the conservatism penalty after a certain point in training.
    Approaches to perform early stopping in offline RL recommend termination based on statistics independent of the loss curves \citep{cql, agarwal2020optimistic}. 
    \item Scenarios occurring in the long tail of a demonstration dataset for a task will appear less frequently when episodes are sampled uniformly from the task environment for model evaluation.
    In this case, computing the validation loss over a uniformly sampled test set or average evaluation reward over uniformly sampled episodes will not accurately reflect the model's progress in resolving causal confusion in this environment. 
\end{enumerate}

To address these issues, we adopt two measures:
First, we evaluate the reward at the end of every epoch (as commonly done in offline RL) and continue training until the reward saturates. 
This allows us to measure both whether active sampling can recover the rewards achieved by uniform sampling in fewer steps and also whether it achieves a higher maximum reward.
Second, we design the evaluation environment episodes in a way that allows isolating the causally confused behaviour and identifying whether causal confusion has been resolved. This amounts to doing a scenario-driven evaluation for the toy environments in which we know what the spurious correlate is by constructing all possible scenarios with and without the spurious correlate present.
This process is described in more concrete terms for the Traffic-World MiniGrid environment in \Cref{traffic-results}.

We will now describe each benchmark environment and discuss the performance of the proposed active sampling schemes in their respective sections.

\subsection{Illustrative Example: Traffic-World}\label{traffic-results}

The autonomous driving literature cites many examples where models trained on large datasets are very performant but exhibit causal confusion on the tail cases of their operational domain, for instance: (1) self-driving agents stopping at pedestrian crossings regardless of whether a pedestrian is present or not since the two often co-occur; (2) agents that simply try to \emph{cruise} if they know their current speed since expert driving datasets contain cruising behaviour in a large fraction of each trajectory.

\begin{figure}[t!]
\begin{minipage}{.39\linewidth}
\includegraphics[width=0.99\textwidth]{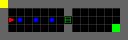}
\\[10pt]
\includegraphics[width=0.99\textwidth]{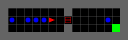}
\end{minipage}
\begin{minipage}{.59\linewidth}
    \includegraphics[width=1\textwidth]{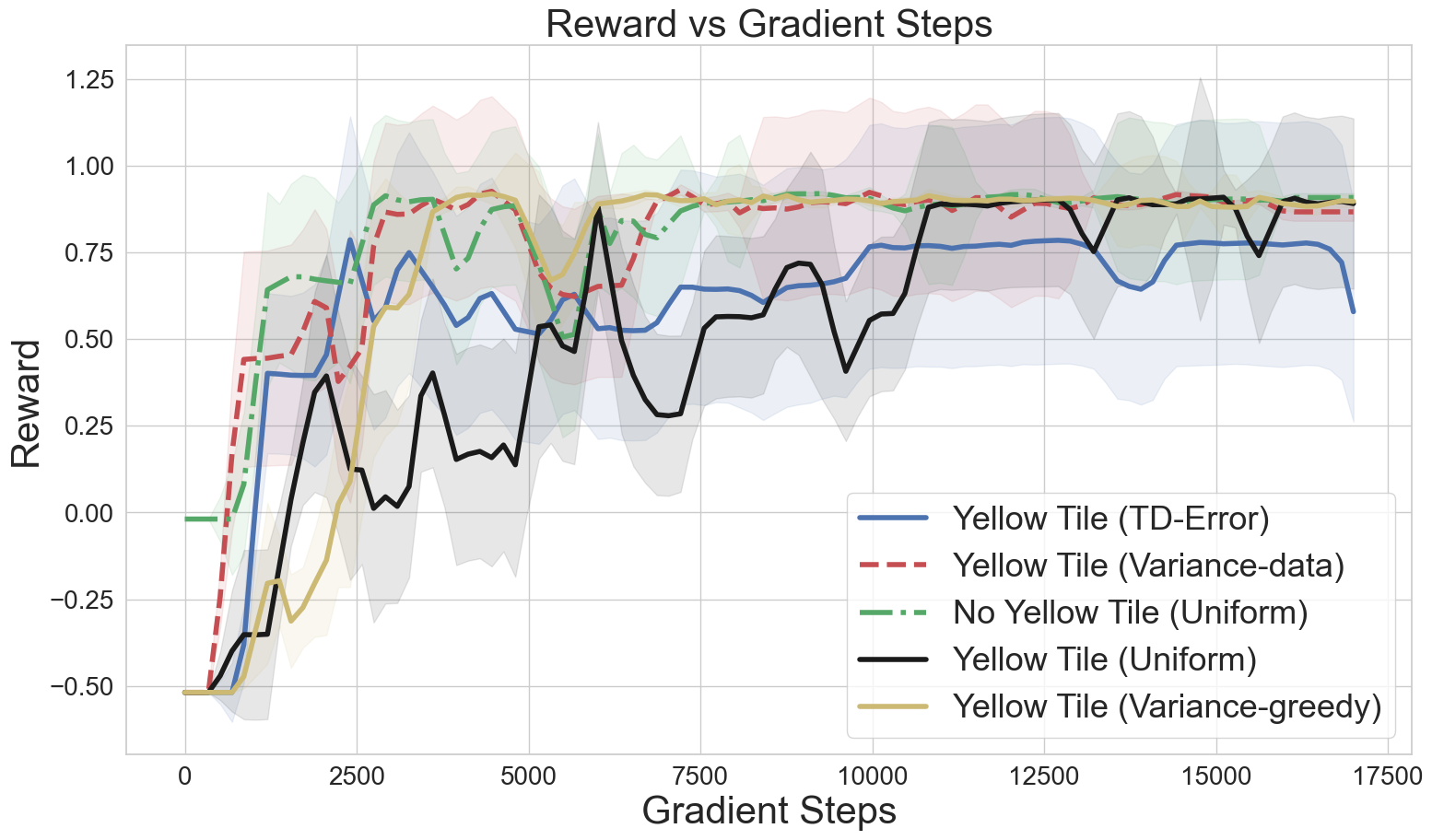}
\end{minipage}

\caption{\textbf{Traffic-World}: \emph{Left Top:} The leading vehicle is static causing the top-left tile to flash yellow. \emph{Left Bottom:} The agent is in front of a red light, and the top-left tile is not yellow since the agent is not blocked by the leading vehicle. \emph{Right:} Reward curves for agents trained on data not exhibiting causal ambiguity  (green) and data exhibiting causal ambiguity (others), i.e. data without and with a spurious correlate respectively. }

  \label{fig:active_vs_random_traffic}
\end{figure}

We build on the traffic gridworld environment proposed in~\citep{clare_cc_rl_explore} (shown in the left panel of~\Cref{fig:active_vs_random_traffic}), where an agent (red triangle) starts at the leftmost point in a row behind randomly initialised leading vehicles (blue circles), and needs to cross a traffic light to reach a goal location (green square) on the right side of the grid. We collect data such that the probability of the traffic light turning red becomes lower as the agent approaches it, and so the data distribution contains:
\textbf{(1)} mostly episodes where the light is green throughout the episode, \textbf{(2)} some episodes where the traffic light is red and the agent has to wait behind the vehicle in front (referenced here onward as the leading vehicle) before the light turns green again, and \textbf{(3)} only a couple of episodes where the light turns red with the agent at the front of the traffic queue.

In this setup, the agent could simply learn to follow the leading vehicle, instead of learning traffic light rules.
To test causal confusion explicitly here, we introduce a related spurious correlate: a flashing yellow tile at the top left of the grid (emulating the brake lights on a leading vehicle), that is yellow whenever the leading vehicle is stopped or blocked, and grey otherwise.
The agent could follow this as an indicator of whether to stop or go ahead, and this policy's actions would be optimal for 98\% of the data points. 

We now define four instances of this environment that constitute our evaluation set, and we average the reward over these when reporting the results.
The first three instances cover tail-cases, and the last one covers the majority case of scenarios that occur in our 7000-episode dataset:
\begin{enumerate}
    \item \textbf{simple-green-with-tile}: The agent is unobstructed; the traffic light is green; the tile flashes yellow. This tests whether the agent relies on the tile to decide when to move.
    \item \textbf{simple-red-no-tile}: The agent is unobstructed; the traffic light is red; the tile does not flash yellow. This tests whether the agent relies on the tile to decide when to stop.
    \item \textbf{traffic-light-switches-with-tile}: The agent starts behind traffic; the traffic light switches to red for some time steps when the agent reaches it; the tile flashes according to the leading vehicle.
    \item \textbf{always-green-with-tile}: This is a trivial episode with traffic where the agent can simply reach the goal without needing to navigate a red light since it is always green.
\end{enumerate}

\Cref{fig:active_vs_random_traffic} shows the evaluation curves of CQL agents trained with uniformly-sampled data, with and without the yellow tile present in images in the dataset. 
We see that the performance of the former agent degrades and it takes ~4$\times$ the number of gradient steps to converge to the solution of the latter agent which is trained without the spurious correlate present. Also shown are the active sampling variants (\emph{TD-Error}, \emph{Variance-greedy} and \emph{Variance-data}) trained with the spurious yellow tile, which perform very similarly to uniform sampling without the spurious correlate present. From the plots, we see that for the TD-error variant there is high variance across seeds; thus the inter-quartile mean of the average rewards across seeds is lower than the maximum achievable reward, although good solutions were found for some random seeds very early in training. 
We also see that the Variance-based versions recover a good solution quickly.

\begin{figure}[t!]
{%
      \includegraphics[width=0.5\textwidth]{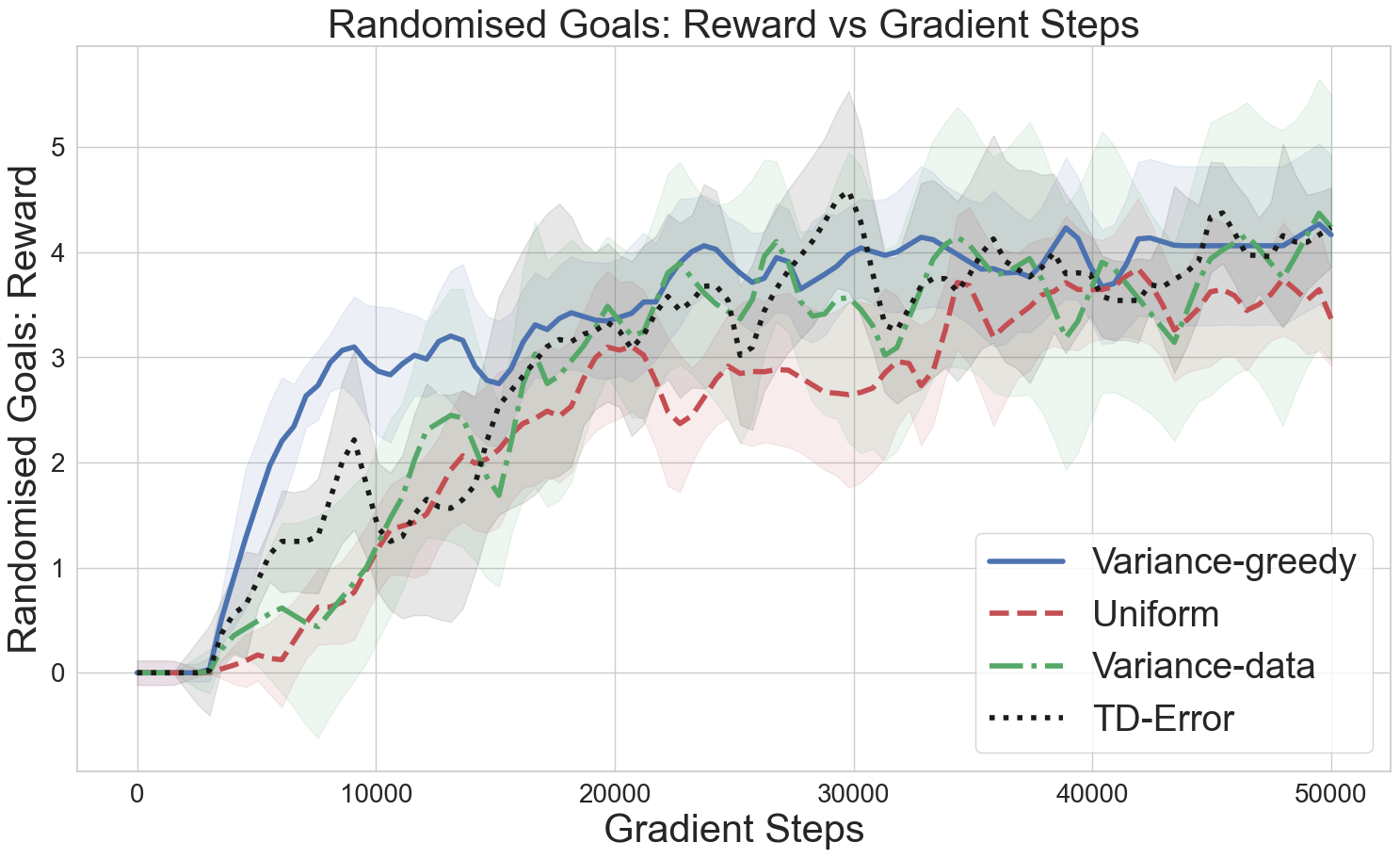}
      \includegraphics[width=0.5\textwidth]{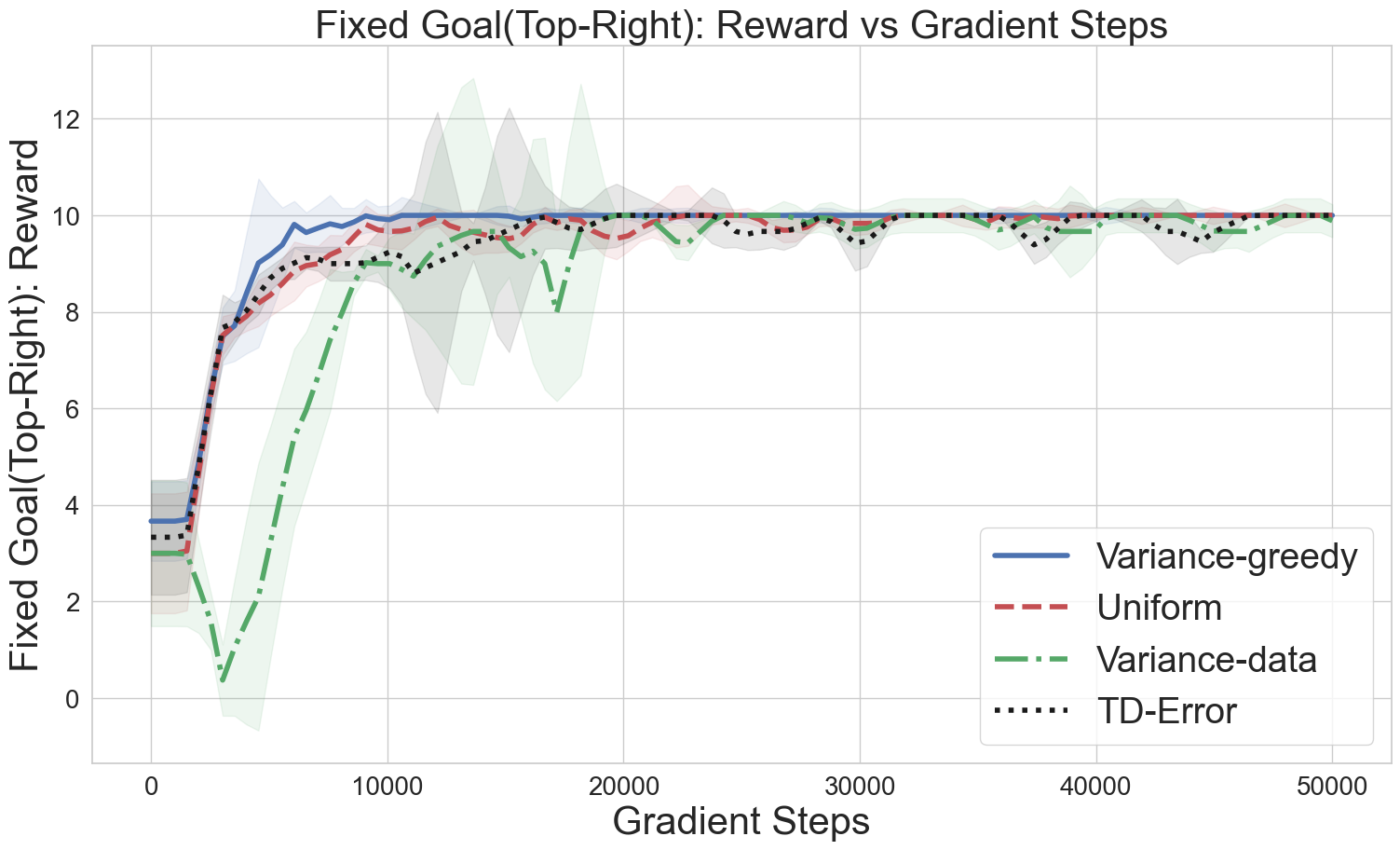}
\caption{(\textbf{Maze}) Rewards curves of different agents trained on a skewed dataset containing 6000 episodes with a fixed goal at the top-right and 200 episodes with a randomly sampled goal in Maze. We see that the agents trained with uniform sampling and active sampling perform similarly on the fixed goal evaluation environment (right), but the active sampling variants achieve a higher reward in the environments with randomly sampled goals (left). This verifies that the model is not just performing well in one of the two kinds of environments and the reason behind the lower performance of uniform sampling, in this case, is causal confusion about the location of the goal.}
\label{fig:perf_norand}
}\end{figure}

\subsection{Assessing Generalization in Offline Reinforcement Learning: Procgen}\label{procgen-results}

The Maze environment in Procgen~\citep{cobbe2019procgen} defines a navigation task where the agent starts at the bottom left in the maze and receives a reward of +10 upon successfully reaching the goal which is sampled at any valid location in the maze.
\citet{goalmisgen} recently showed that an agent trained on a series of environments with the goal always at the top-right will be causally confused about the source of the reward and will still navigate to the top-right even when the goal is sampled elsewhere. 

We generate a skewed \emph{mixture} dataset containing mostly episodes where the goal is sampled at the top-right, and a few episodes where the goal is sampled randomly.
Further details about the collection are described in the Appendix.
\Cref{fig:perf_norand} shows the evaluation performance of random and active sampling agents trained on this \emph{mixture} dataset.
The left and right plots show the performance when the goals are sampled randomly and from the top-right in the evaluation environment, respectively.
We observe that active sampling recovers the maximum performance achieved by uniform sampling in half the number of training steps.
On this benchmarking task, the gains from active sampling are largely in terms of training efficiency since both active and uniform sampling variants saturate to similar final rewards in the environment with randomly sampled goals.
We plot the computation time for the uniform and active sampling variants in~\Cref{fig:timing_procgen} in the appendix and note that variance-based sampling reaches the highest-score achieved by uniform sampling in lesser wall-clock time.
Qualitative evaluations show that uniform sampling agents which achieve a lower reward still successfully navigate to the top-right corner of \emph{hard} mazes, and are therefore only confused about the location of the goal. 

It is important to note  that if the evaluation environments considered here had followed a distribution of scenarios similar to the training data distribution, then we would have observed saturation of the evaluation reward somewhere between 10,000--20,000 gradient steps.
This is because the evaluation performance on the environments with the fixed goal, which occurs much more frequently, saturates at this point. 
An active sampling model on which early-stopping would be performed based on the validation loss, would then have fared much better on the tail case of randomly-sampled-goal episodes (and done similarly well otherwise), as compared to the model trained with uniform sampling of data.
This shows why the experimental design choices we adopted in this work---scenario-driven evaluation and termination of training based on rewards---are crucial to appropriately evaluating causal confusion in offline RL agents.

\begin{figure}[t!]
{%
      \includegraphics[width=0.5\textwidth]{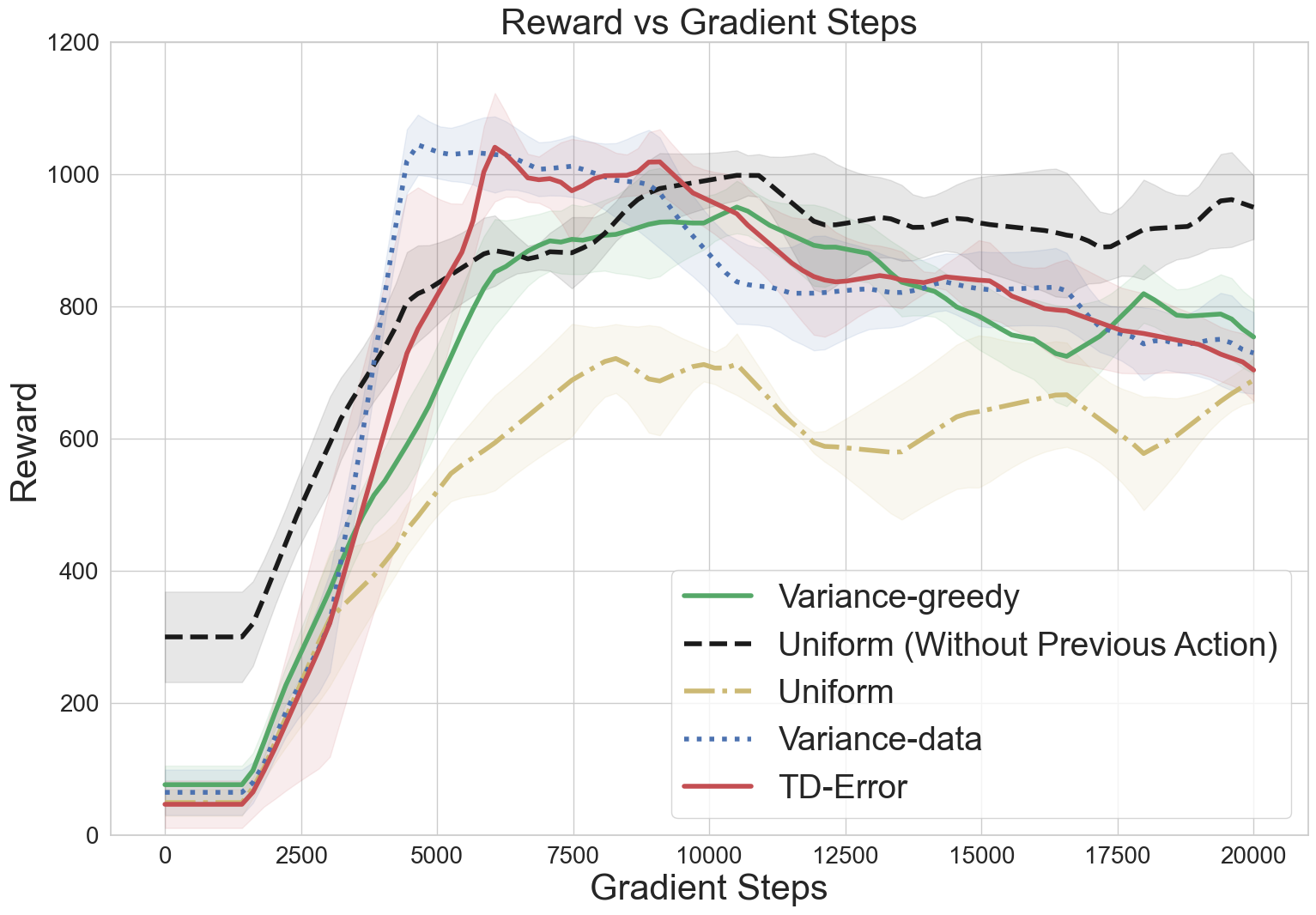}
      \includegraphics[width=0.5\textwidth]{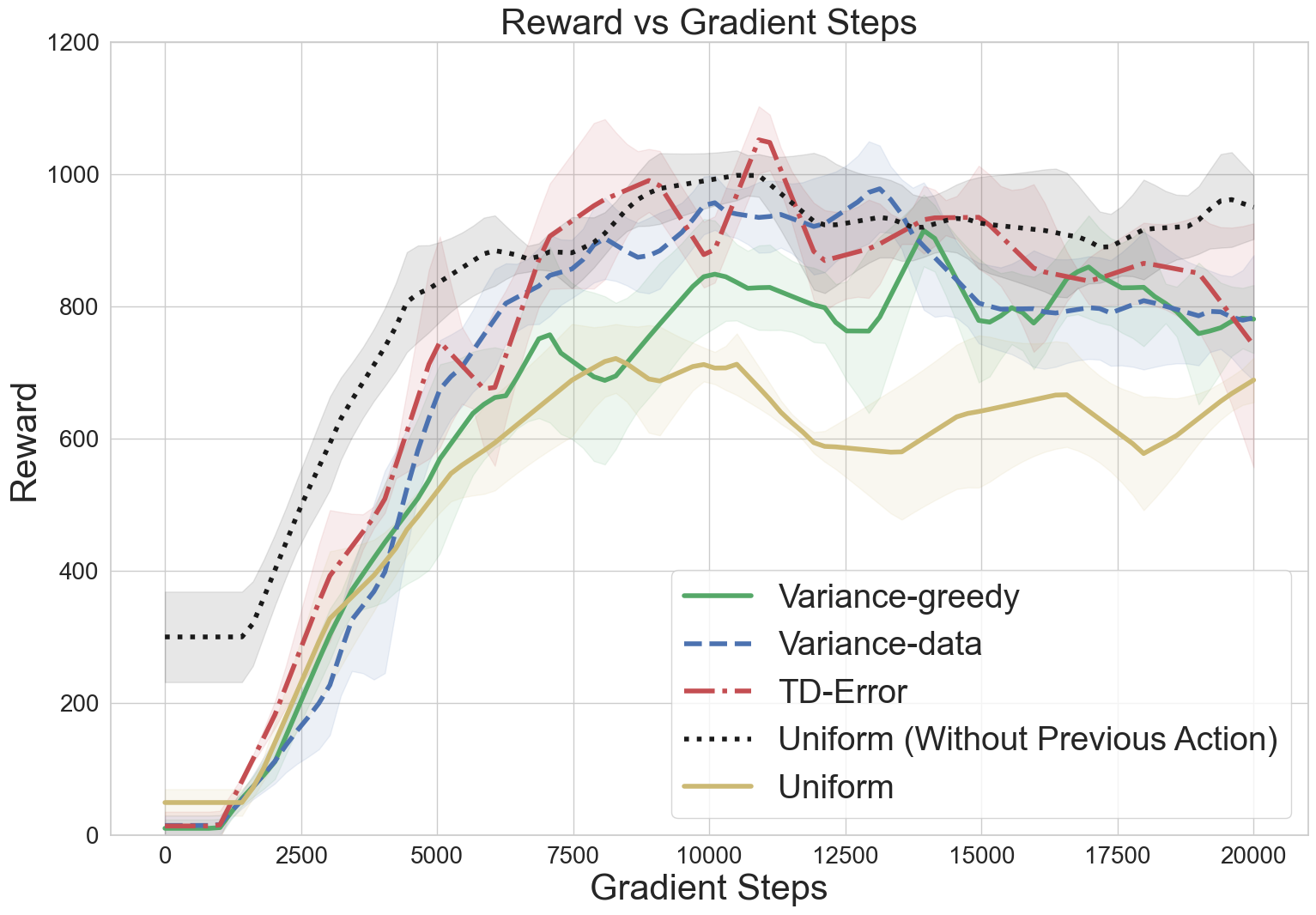}
\caption{Comparison of active and uniform sampling agents on Enduro. On the left is the comparison in the \emph{-dataset} case, and on the right is the case with \emph{-batch} sampling where scores are incremented and updated based on the sampled batch. We see an increase in the number of gradient steps it takes the \emph{-batch} case to reach the highest possible reward.}
\label{fig:perf_dataset}
}
\vspace{-10pt}
\end{figure}

\vspace*{-5pt}
\subsection{Causal Confusion in the ALE Benchmark} \label{ale-results}

We are now interested in evaluating our active sampling baselines on a larger pre-existing dataset with more realistic noise and variations.
Prior work in imitation learning ~\citep{oreoneurips,dehaan} has attempted to simulate causal confusion in specific Atari game-play datasets by modifying images to display the previous action taken by the agent. 
This kind of causal confusion is inspired by robotics datasets that have trajectories with highly correlated (and thus predictable) sequences of actions, because embodiment dictates that an agent's state does not change too drastically between subsequent timesteps.

\begin{figure}[t!]
\begin{minipage}{.5\linewidth}
      \includegraphics[width=1\textwidth]{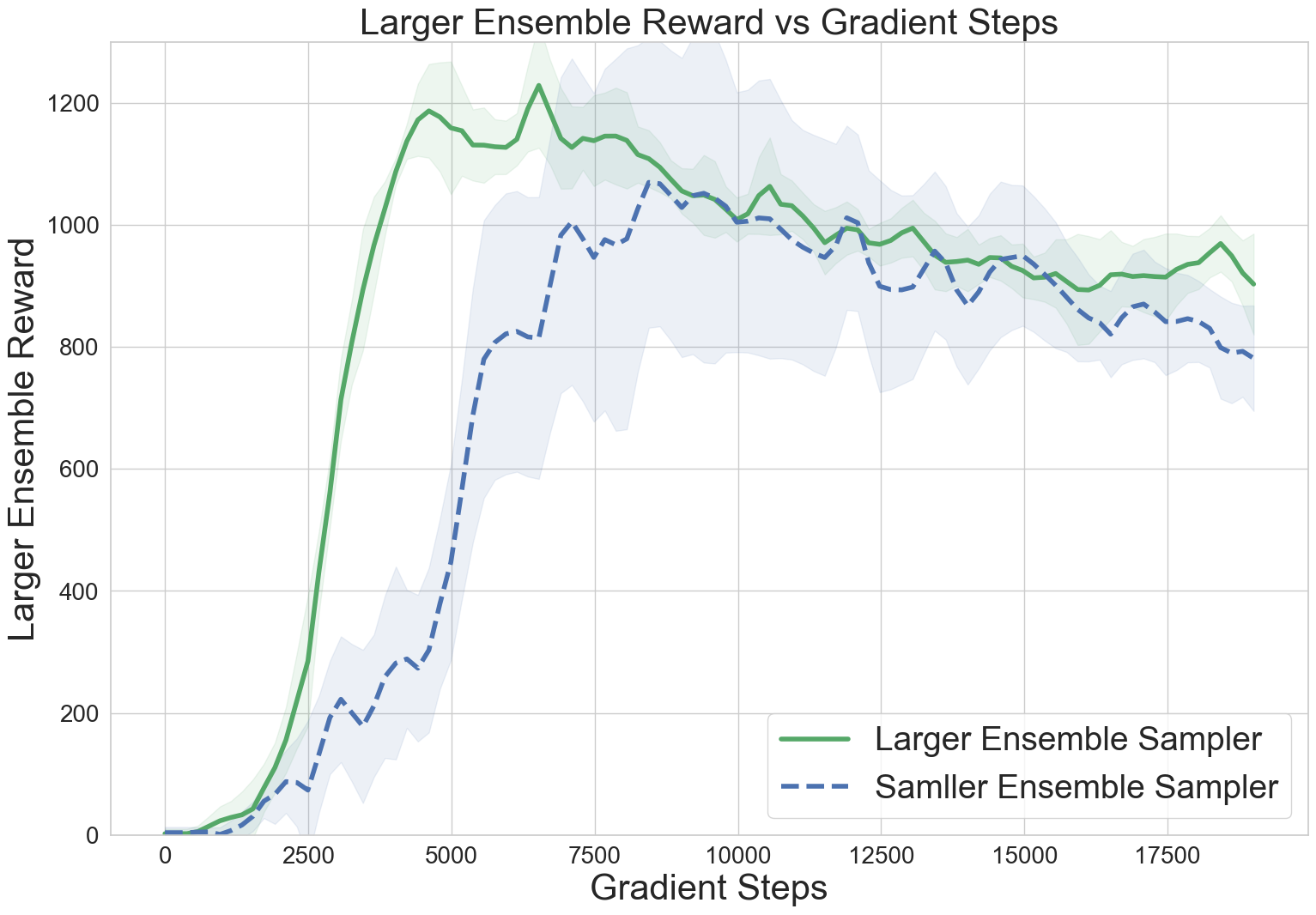}
\end{minipage}
\begin{minipage}{.5\linewidth}
      \includegraphics[width=1\textwidth]{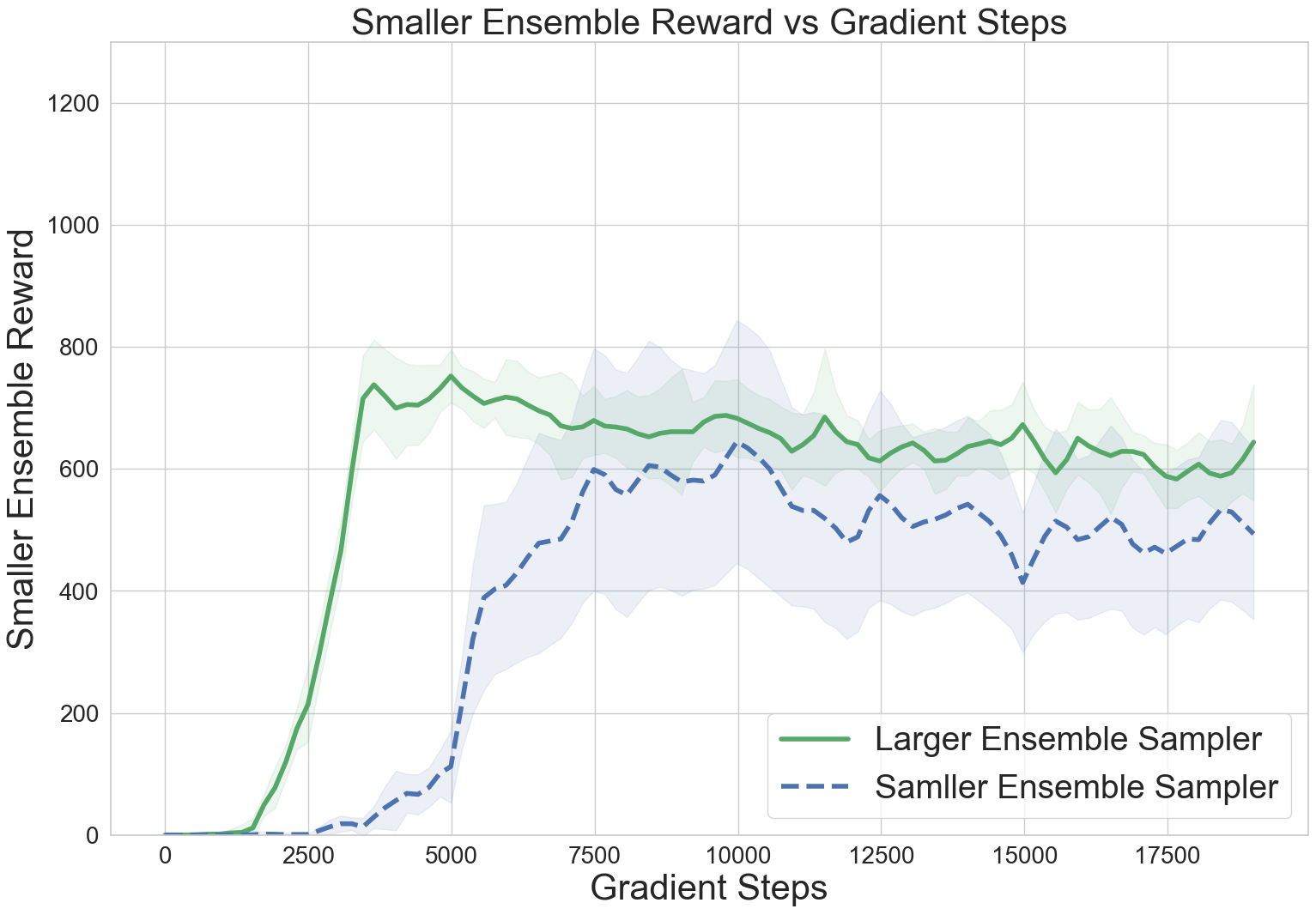}
\end{minipage}
\caption{Reward curves of the larger (left) and smaller (right) ensemble training when sampling on the basis of the uncertainty of either one: of the speed of convergence depending on the quality of uncertainty estimation.}
  
\label{fig:perf_larger_sampler}
\end{figure}

\begin{table}[h!]
\centering
 \begin{tabular}{c|c|c|c|c} 
 \hline
 \hline
 \multicolumn{1}{p{3cm}}{\centering Fraction of  \\ data sub-sampled }
 & \multicolumn{1}{|p{2cm}|}{\centering Reward \\ (Uniform) }
& \multicolumn{1}{|p{3cm}|}{\centering Reward \\ (Variance-data)}
& \multicolumn{1}{|p{3cm}|}{\centering Reward \\ (Variance-greedy) }
& \multicolumn{1}{p{2cm}}{\centering Reward \\ (TD-Error) }\\
 \hline
 0.5 &  470 &  780   & 560  &   560  \\ 
 0.7 &  570  &  850  & 810  &   820  \\
 0.9 &  660 &  1000  & 890  &   980  \\ [1ex] 
 \hline
 \hline
 \end{tabular}
 \caption{Sub-sampling experiment on the Enduro dataset.}
 \label{table:enduro_subsampling}
\end{table}

However, when we trained uniform sampling CQL agents with and without the previous action on the images for 20 Atari games, only a select few games actually exhibited causal confusion of this kind.
We pick Enduro since we can consistently observe both a convergence speed degradation and final reward degradation upon adding the previous action to the image.
Enduro is a car racing game where an agent needs to overtake cars (receiving a positive reward for each car overtaken) and drive along a winding road, with a limited number of collisions allowed.

\Cref{fig:perf_dataset} shows our proposed active sampling baselines compared to uniform sampling in the following two cases that we previously discussed in \Cref{computing}: (1) \emph{-dataset} case when we recompute the scores across the dataset every few gradient steps and (2) \emph{-batch} case when the scores are only recomputed for the sampled batch and updated in a priority queue.

Also shown is the uniform sampling baseline trained without the previous action displayed on the image (data without causal ambiguity).
We note that all active sampling variants perform significantly better than their uniform sampling counterpart when both are trained with the spurious correlate (i.e., with the previous action displayed on the image).
This discrepancy is observed in terms of convergence time to their best performance and in terms of the highest reward achieved, which is approximately 40 percent higher than the reward achieved by uniform sampling.
While both \emph{TD-Error} and \emph{Variance}-based variants achieve a similar highest score, \emph{TD-error} sampling takes twice the number of forward passes (and thus twice as long) to compute the acquisition score (i.e., the TD-error) in the \emph{-dataset} case.

We further observe that when we go from the \emph{-dataset} to the \emph{-batch} setting (i.e., updating scores only for the sampled batch, \Cref{fig:perf_dataset}, right), the final reward achieved by the active variants is only slightly lower than previously.
However, in this setting, all methods require twice the number of gradient steps to obtain this reward since the acquisition scores in the \emph{-batch} setting are stale for many data points.
A relevant line of future work will be to investigate how to maintain the sampling quality without needing to recompute all scores as in the \emph{-dataset} case.

We also conducted an experiment where we uniformly sub-sample decreasing fractions of the original dataset and see that even the active sampling variants trained with 70\% of the data outperform the uniform sampling variants trained on all the data.
This suggests that naively acquiring more data is unlikely to resolve causal confusion if the data distribution remains skewed.
We present these results in \Cref{table:enduro_subsampling}.

\subsection{Effect of Predictive Uncertainty on Active Sampling and Causal Confusion} \label{uncertainty-quality-results}

To better understand the improvement in convergence speed of active sampling vis-`a-vis uniform sampling, we train CQL with a smaller ($n=3$) and a larger ($n=10$) ensemble of $Q$-networks, all else equal, and sample identical transitions based on the predictive uncertainty of only either the smaller or the larger ensemble.
This experiment setup allows us to isolate the effect of the predictive variance on active sampling and therefore on the observed reduction in causal confusion.

\Cref{fig:perf_larger_sampler} shows the training curves for both ensembles, and we find that when transitions are sampled according to the predictive uncertainty of the larger ensemble, even the smaller ensemble converges faster (albeit to a smaller value than that achieved by the larger ensemble).
Similarly, when points are sampled according to the predictive uncertainty of the smaller ensemble, the larger ensemble converges to its highest reward more slowly.
This result implies that improved predictive uncertainty estimation, for which we use the size of the ensemble as a proxy, improves the ability of active sampling to identify samples that break spurious correlations in the data and reduce causal confusion.
We provide a further discussion of these observations in~\Cref{further}.

\vspace*{-5pt}
\section{Conclusion}

In this paper, we studied how to alleviate causal confusion in offline RL.
We designed uncertainty- and loss-based data sampling baselines to selectively sample transitions for training, and found evidence that active sampling can recover a less causally-confused model in significantly fewer training steps as compared to uniform sampling.
In future work, we hope to scale the analysis performed in this paper to larger benchmark environments, with sources of noise in the reward and transitions, to further corroborate our findings.
Such an analysis would help further distinguish the quality of solutions found through loss-based and uncertainty-based active sampling since noisy transitions can have \emph{irreducible} loss.
Additional promising avenues for future work include studying the usefulness of active sampling in the regime of offline-to-online RL-based fine-tuning, as well as extending the analysis presented in this paper to environments with continuous action spaces.
\vspace{-4pt}
\section*{Acknowledgements}

We thank anonymous reviewers for useful and constructive feedback on this manuscript.
We also thank Clare Lyle, Karmesh Yadav, and Cong Lu for useful discussions and for assistance in setting up the benchmark environments.
Gunshi Gupta is funded by the EPSRC Centre for Doctoral Training in Autonomous Intelligent Machines and Systems (EP/S024050/1) and Toyota Europe.
Tim G. J. Rudner is funded by the Rhodes Trust, the Engineering and Physical Sciences Research Council (EPSRC), and a Qualcomm Innovation Fellowship.
We gratefully acknowledge donations of computing resources by the Alan Turing Institute.

\bibliography{references}

\clearpage

\begin{appendices}

\crefalias{section}{appsec}
\crefalias{subsection}{appsec}
\crefalias{subsubsection}{appsec}

\setcounter{equation}{0}
\renewcommand{\theequation}{\thesection.\arabic{equation}}

\onecolumn

\vspace*{-20pt}
{\hrule height 1mm}

\section*{\LARGE \centering Supplementary Material}
\label{sec:appendix}

\vspace{5pt}
{\hrule height 0.3mm}
\vspace{14pt}

\vspace*{20pt}
\section*{\Large Table of Contents}
\startcontents[sections]
\printcontents[sections]{l}{1}{\setcounter{tocdepth}{2}}
\vspace*{-10pt}

\clearpage

\section{Conditional Average Treatment Effect Estimation from Causally Ambiguous Data} \label{cate-connections}
In this section, we discuss a causal interpretation of our approach which employs active sampling as a tool for causal discovery.
We start by describing quantities of interest in the field of causal inference and their connection to $Q$-values (and derived quantities) used in RL.

Treatment-effect estimation, where the goal is to estimate the effect of a treatment ${T} \in \mathcal{T}$ on the outcome ${Y} \in \mathcal{Y}$ for individuals described by covariates ${X} \in \mathcal{X}$, is a central problem in causal inference.
In particular, we may wish to estimate the expected difference in potential outcomes for individuals when subjected ($t=1$) or not subjected ($t=0$) to a treatment $t$, measured by the Conditional Average Treatment Effect (CATE,~\citet{cate}), defined as
\begin{align}
    \tau({X}) \equiv \mathbb{E}[Y \mid x, t=1]-\mathbb{E}[Y \mid x, t=0]      .
\end{align}
Here the treatment $t$ is often considered to be a binary variable, but the definition can be extended to the multivariate case with continuous or discrete values.
Realizations of the random variables $X, T, Y$ are denoted by $x, t, y$, respectively. 
The set of assumptions needed to ensure identifiability of the CATE estimator are listed in~\citep{Rubin1974EstimatingCE}.

To frame the problem of disambiguating the effect of different actions (i.e., treatments $t \in T$) in a given state (i.e., a set of covariates $x \in \mathcal{X}$), we frame estimation of the advantage functions $A(s, a) = Q(s, a) - V(s)$ in reinforcement learning as CATE estimation~\citep{dae}.
In particular, the outcome $Y$ corresponds to the $Q$-function (i.e., the expected return) for a given state--action pair, and we can express the corresponding CATE estimator as
\begin{align}
    \mathbb{E}\left[\sum_{t=0}^{\infty} \gamma^t r\left(\mathbf{s}_t, \mathbf{a}_t\right) \mid s_t=s, a_t=a\right]-\mathbb{E}\left[\sum_{t=0}^{\infty} \gamma^t r\left(\mathbf{s}_t, \mathbf{a}_t\right) \mid s_t=s\right]=Q(s, a)-V(s)=A(s, a),
\end{align}
which indicates the advantage of executing an action $a$ at state $s$ as opposed to any other alternative action.
With this connection established, we can now cast active sampling of data points to estimate a value or advantage function, as a sequential process of accurately estimating treatment effects.

\section{Evaluation Metrics}
We present the reward curves directly in most of our experimental reporting. The curves are computed by taking the inter-quartile-mean (IQM) across seeds as proposed by \citet{statisticalrl}. We report the number of seeds used for all experiments in~\Cref{hyperparam}.

Offline RL training performance is known to be non-monotonic, unlike supervised learning where the accuracy (or loss) increases (decreases) and then saturates. Often the reward curves start to decrease after a period when overfitting to the data-action values (through the conservatism penalty) starts to happen. Since we define deterministic benchmark environments testing all the tail-scenarios in the case of Traffic-World and Maze experiments, we can consider any point on the curve where the model solves the highest number of environments (achieves the highest reward) as the point of convergence (as opposed to it being a noisy spike due to stochastic evaluation). This is similar to taking the max-reward checkpoint as done in \citep{pmlr-v119-agarwal20c}. However, to ensure that the solutions learned by any method are recoverable, we want to be able to get a good checkpoint from the model without needing to evaluate it too often. Thus we consider a method as having achieved higher reward than other methods at any point in time if it maintains this gap for at least two subsequent post-epoch evaluations. This is akin to taking a windowed-max over the reward curves.

\clearpage

\section{Implementation}
All our environments use a discrete action space. Therefore we build our method on top of the double-DQN implementation similar to the original CQL paper. As stated in~\Cref{eqn:cql_objective}, we use ensembles of $Q$-networks, and at evaluation time, we average the $Q$-value outputs of the ensemble, and select the action with the maximum $Q$-value.
In other places where we need to do inference (for instance: to compute $Q$-values for the conservative loss) we take the mean across the ensemble.

A design choice we make is the network initialisation when we start to do active sampling: the uncertainty encoded by a random network at the start of training can be very inaccurate and biased towards some subset of the data. We instead train with uniform sampling for an epoch and then start active sampling second epoch onwards. As expected, we see that a partially trained ensemble encodes better uncertainty compared to a randomly initialised one and that bigger ensembles encode better uncertainty than smaller ones when the network is not trained enough with uniform sampling at the start, as seen in \Cref{fig:perf_larger_sampler_longer_random-training}.

\section{Further Analysis} \label{further}

\begin{wrapfigure}{r}{0.5\textwidth} 
\vspace*{-45pt}
  \begin{center}
    \includegraphics[width=0.5\textwidth]{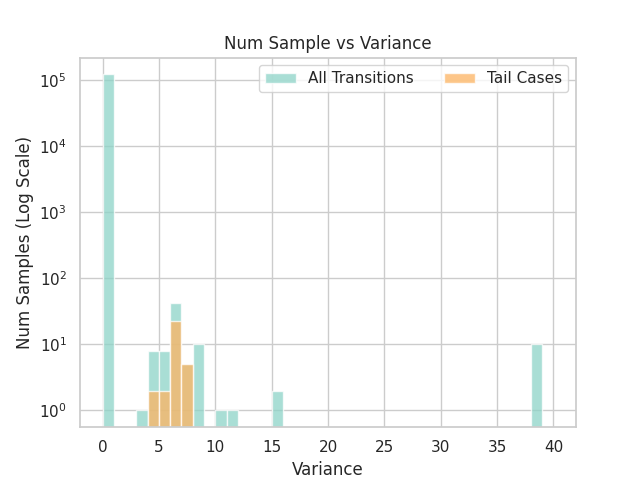}
    \caption{(\textbf{Traffic-World}) Histogram of the advantage function variance estimates for the data action plotted for all transitions, overlayed by the scores for one tail case's samples. The tail case we consider here comprises state-action pairs where the top-left tile flashes yellow but the data-collecting policy still moves one step ahead ignoring the tile, since there is space to move towards the goal without getting a negative penalty or triggering an episode termination. We can see that these samples fall in the 98$^\text{th}$ percentile of all points.}
    \label{fig:histo}
  \end{center}
\end{wrapfigure}

\textbf{Uncertainty Histogram:} In \Cref{fig:histo}, we plot a histogram for the computed \emph{Variance-data} scores after one epoch of training with uniform sampling on the Traffic-World benchmark. We overlay the scores for a specific tail case over the scores for the entire dataset. We can verify that samples from the tail-case fall in the higher percentile of variance scores.

\begin{figure}
  {%
      \includegraphics[width=0.5\textwidth]{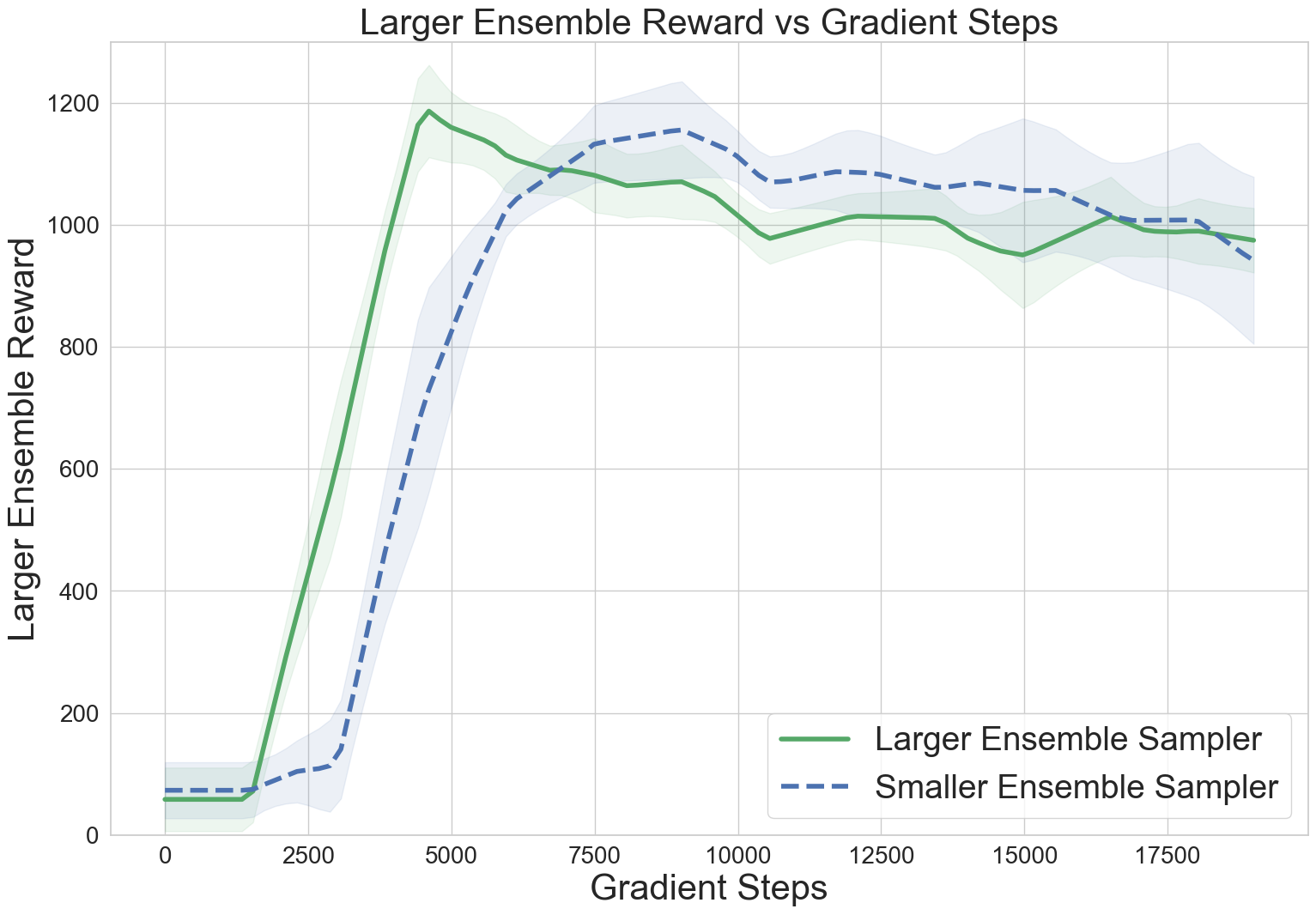}
      \includegraphics[width=0.5\textwidth]{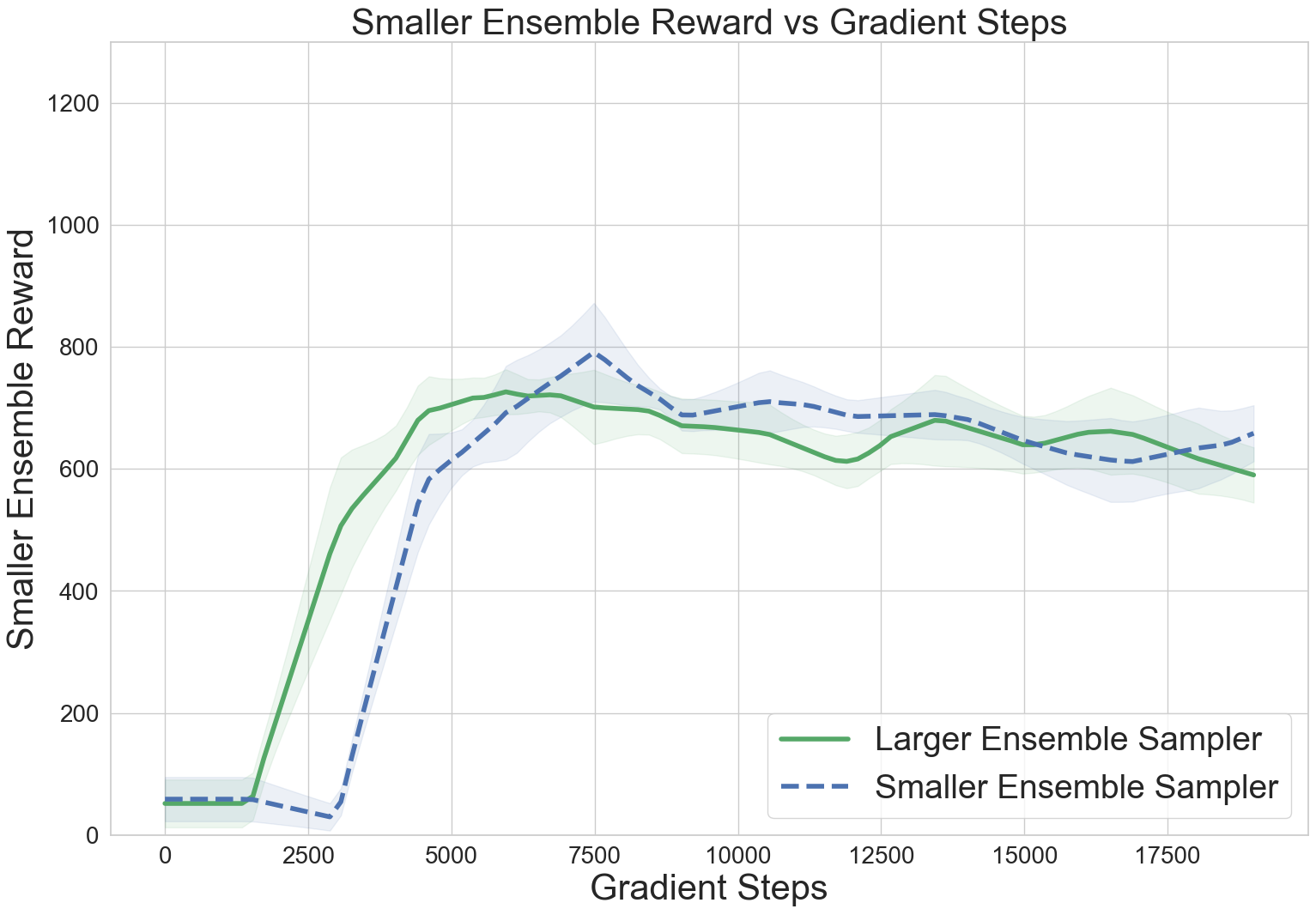}
    \caption{Reward curves of the larger (left) and smaller (right) ensemble training when sampling on the basis of the uncertainty of either one: of the speed of convergence depending on the quality of uncertainty estimation. Here, the difference is that both ensembles start active sampling from an initialisation that is trained for longer with uniform sampling. We see a less drastic gap in convergence between the two samplers, as compared to the main paper's~\Cref{fig:perf_larger_sampler}. This is expected since the uncertainties should get more informative (up to a point) as we train for longer with random sampling.}
\label{fig:perf_larger_sampler_longer_random-training}%
\vspace*{-10pt}
}\end{figure}

\textbf{Ensemble Uncertainty over a longer period of training:} In addition to the figure in the main paper analysing the quality of uncertainty estimation on the Atari benchmark, we plot in~\Cref{fig:perf_larger_sampler_longer_random-training} the training curves of two ensembles when they've been initialised with a longer period of uniform sampling at the start. We see that the gap between the informative-ness of samples fetched based on their uncertainties reduces (as compared to~\Cref{fig:perf_larger_sampler}). The larger ensemble's uncertainty still leads to quicker convergence, however, than when both ensembles are trained using the smaller ensemble's uncertainty for data-sampling.

\begin{wrapfigure}{r}{0.5\textwidth} 
  \begin{center}
    \includegraphics[width=0.5\textwidth]{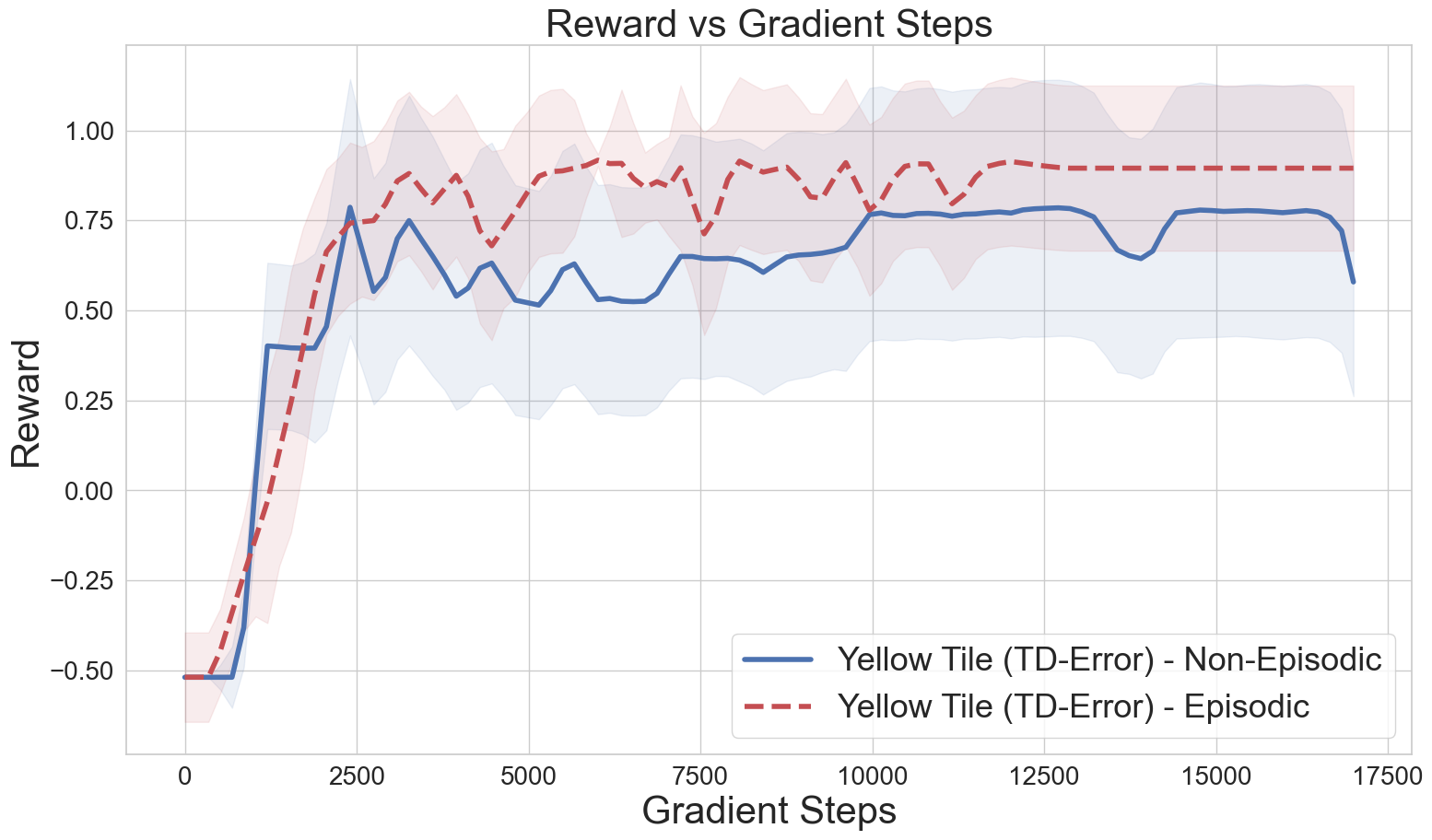}
    \caption{(\textbf{Traffic-World}) Two agents trained with and without episodic scores for TD-Error-based active sampling of causally ambigous data.}
    \label{fig:timing_traffic}
  \end{center}
\end{wrapfigure}

\textbf{$Q$-value divergence and Episodic Sampling to ensure better propagation of TD-Error:} Another observation we made was related to active sampling results in the case of benchmarks with short episodes ($n_{steps} \sim 20-30$) and sparse rewards, in our case Traffic-World and Maze. Q-learning is trained through bootstrapping where we minimise the TD-error which involves estimating the $Q$-values at successive states across transitions in a trajectory. Additionally, the $CQL$ objective has the gap-expanding property because the conservatism penalty tries to push $Q$-values of different actions at a state apart (pushing up the data-action value and pushing down others to some extent).
Sometimes in the case of repeated sampling of a particular transition tuple, there is a potential divergence of $Q$-values of the nearby (preceding and following) state-action pairs. This can be seen as an explosion of $Q$-values in the training metrics and can be partially resolved by gradient clipping. We only observed this in the case of $TD-Error$ and $Variance-data$ variants.

One experiment we tried to address the above concern (with some success) was that of \emph{episodic} sampling, where we use a heuristic to convert individual transition-wise acquisition scores to scores over entire episodes (for example taking the maximum acquisition score over transitions in an episode). This kind of episodic sampling turns out to give much more stable training curves but involves additional hyperparameters and heuristics. The reasoning for it's success is likely related to the motivation of algorithms like emphatic-TD \citep{emphatic}, Reverse Experience Replay (RER) \citep{reverse}. In these works, they propose not just prioritising transitions with high TD-error, but also increasing the priority of transitions preceding these ones in the priority queue, since these transitions will have informative TD-updates in the subsequent time-steps. We show the reward curves with episodic and without episodic active sampling of data corresponding to the experiment in Traffic-World with the spurious correlate present, in \Cref{fig:timing_traffic}.

\begin{wrapfigure}{r}{0.5\textwidth} 
\vspace{-30pt}
  \begin{center}
    \includegraphics[width=0.45\textwidth]{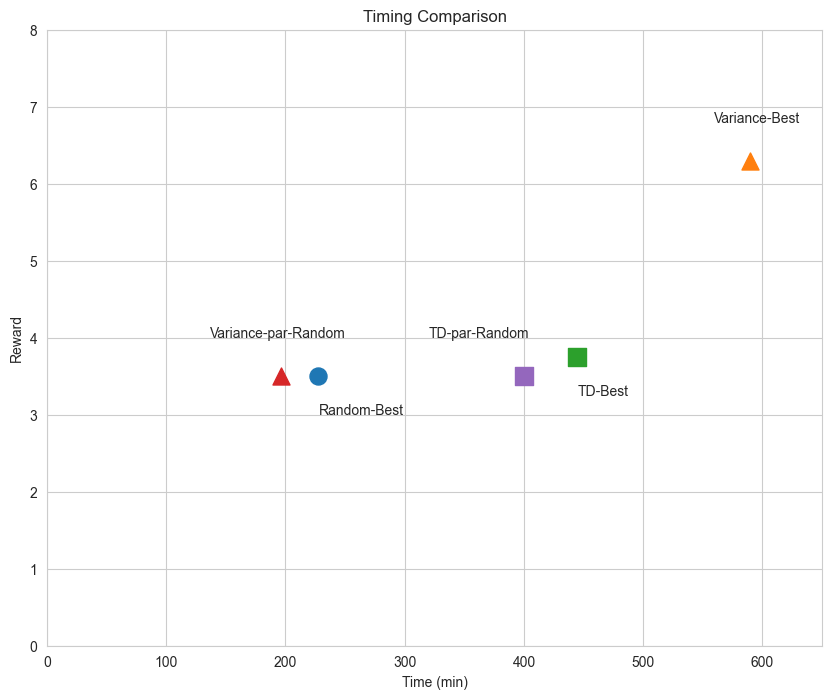}
    \caption{Timing Comparison for different sampling schemes on the Procgen-Maze benchmark plotted as reward achieved versus wallclock time in minutes. }
    \label{fig:timing_procgen}
  \end{center}
  \vspace*{-35pt}
\end{wrapfigure}

\section{Computational Cost}

\Cref{fig:timing_procgen} shows a scatter plot for the wall-clock times to achieve highest reward across different active and uniform baselines (labelled as TD-Best, Variance-Best and Random-Best) corresponding to the results for the Maze environment.
It also plots the time needed for active sampling variants to achieve the best reward that uniform sampling achieves (denoted as Variance-par-Random and TD-par-Random in~\Cref{fig:timing_procgen}).
We note that we use a very simple implementation of active sampling in our work which can be optimised significantly by paralleling the computation of acquisition scores over the dataset. This will make the trade-off between performance and training time even more favourable towards approaches that do active sampling.

\vspace*{-5pt}
\section{Data Collection}

\textbf{Traffic-World.}
To collect data for Offline RL, we trained a PPO agent on a slightly modified version of the Traffic-world environment, with reward shaping on the environment \citep{clare_cc_rl_explore}, to incentivise the agent to reach the goal since this is a hard exploration environment (there is the potential to receive many negative rewards before receiving a positive reward, and without reward shaping the PPO agent just learns to toggle in-place till the episode ends to avoid negative penalties). 
\textbf{Maze.}
We use the Impala-based PPO agent trained in~\citep{goalmisgen} for 200M steps to collect the expert trajectories on 6000 episodes of episodes with randomised goals and 200 episodes of episodes with fixed goals.

\vspace*{-5pt}
\section{Hyperparameters} \label{hyperparam}

For all benchmarks, we performed a grid-search around the hyperparameters from CQL \citep{cql}.
We found the combination of parameters that matches previously reported scores for the Procgen and Atari benchmarks and achieves the highest score on the Traffic-World benchmark, when using uniform sampling with CQL.
For active sampling, we used the same hyperparameters as for uniform sampling (batch size, learning rate, $\alpha$).
Additional parameters related to active sampling are: \textbf{(1)} $n$: the number of gradient steps we take before we recompute acquisition scores on the data. Thus, $n-1$ is the number of gradient steps for which the scores remain stale.
\textbf{(2)} the ensemble size which we keep constant across the active and active sampling variants for a fair comparison.

For the \emph{-batch} case of incrementally computed scores, the hyper-parameters involved are temperature and increment coefficients for importance weighting of gradients ($\beta$ and $inc$), the temperature $\alpha$ for the acquisition scores and epsilon $\epsilon$ (the additive constant for the scores to ensure no data-point's priority is equal to zero and thus never sampled). We used the same values as those used in the original PER implementation \citep{per}, and set $\beta$ and $inc$ to zero, as we found training to work better without importance weighting of gradients. \Cref{table:hyper_traffic,table:hyper_procgen,table:hyper_atari} list out the final values chosen for reporting results across the three benchmarks, along with the values used for the grid search.

\clearpage

\begin{table}[t!]
\centering
\begin{tabular}{l|l|l} 
\centering
Hyperparameter & Value & Search \\
\hline $\alpha$ (CQL) & 1 & 1, 4\\
learning rate & $5 \times 10^{-3}$ & $5 \times 10^{-4}$ , $1 \times 10^{-3}$ , $5 \times 10^{-3}$ \\
batch-size & 512 & 256, 512, 1024\\
$n$ (steps before score recomputation) & 4  & 2,4,8,16 \\
gradient clipping norm & 5 & 1,3,5,7\\
target update interval & 4 & 1, 4, 16, 32\\
ensemble size & 3 & 3,6\\
seeds & 7 & \\
\label{tab:traffic}
\end{tabular}
\caption{Traffic-World experiments.}
\label{table:hyper_traffic} 
\end{table}

\begin{table}[t!]
\centering
\begin{tabular}{l|l|l} 
\centering
Hyperparameter & Value & Search \\
\hline $\alpha$ (CQL) & 1 & 1, 4\\
learning rate & $1 \times 10^{-3}$ & $5 \times 10^{-4}$ , $1 \times 10^{-3}$ , $5 \times 10^{-3}$\\
batch-size & 2048 & 1024, 2048\\
$n$ (steps before score recomputation) & 8  & 4, 8, 16\\
gradient clipping norm & 5 & 1,3,5,7\\
target update interval & 50 & 20, 50, 100\\
ensemble size & 5 & 3, 5\\
seeds & 9 & \\
\label{tab:maze}
\end{tabular}
\caption{Maze experiments.}
\label{table:hyper_procgen} 
\end{table}

\begin{table}[t!]
\centering
\begin{tabular}{l|l|l} 
\centering
Hyperparameter & Value & Search \\
\hline $\alpha$ (CQL) & 1 & 1, 4\\
learning rate & $5 \times 10^{-3}$ & $5 \times 10^{-4}$ , $1 \times 10^{-3}$ , $5 \times 10^{-3}$ \\
batch-size & 2048 & 1024, 2048\\
$n$ (steps before score recomputation) & 8  & 8, 16\\
gradient clipping norm & 7 & 1,3,5,7 \\
target update interval & 100 & 10, 50, 100, 1000\\
ensemble size & 5 & 3, 5, 10\\
seeds & 9 & \\
\label{tab:enduro}
\end{tabular}
\caption{Enduro experiments.}
\label{table:hyper_atari} 
\end{table}

\end{appendices}

\end{document}